\documentclass{article}
\usepackage{spconf,amsmath,graphicx,hyperref}
\usepackage{enumitem}
\usepackage{cite}
\usepackage{amssymb,amsfonts}
\usepackage{multirow}
\usepackage{algorithmic}
\usepackage{textcomp}
\usepackage{xcolor}
\usepackage{colortbl}

\usepackage{booktabs}
\usepackage{float}
\usepackage{multirow}
\usepackage{makecell}

\usepackage{fontawesome5}

\usepackage{bm}
\usepackage{xspace}
\usepackage{footnote}
\usepackage{bbding} 
\makeatletter
\DeclareRobustCommand\onedot{\futurelet\@let@token\@onedot}
\def\@onedot{\ifx\@let@token.\else.\null\fi\xspace}

\def\etal{\emph{et al}\onedot}
\makeatother
\usepackage{epstopdf}

\usepackage{xcolor}
\PassOptionsToPackage{numbers}{natbib}
\definecolor{url}{RGB}{0,73,147}
\definecolor{mypink}{HTML}{bc4749}
\definecolor{mygray}{gray}{.85}

\usepackage{multirow}
\usepackage[table]{xcolor}
\makeatletter
\newcommand{\thickhline}{%
\noalign {\ifnum 0=`}\fi \hrule height 1pt
\futurelet \reserved@a \@xhline
}
\makeatother

\def\BibTeX{{\rm B\kern-.05em{\sc i\kern-.025em b}\kern-.08em
T\kern-.1667em\lower.7ex\hbox{E}\kern-.125emX}}

\usepackage{amsmath,amsfonts,bm}

\def\eqref#1{equation~\ref{#1}}

\def\1{\bm{1}}

\def\mE{{\bm{E}}}

\def\mP{{\bm{P}}}

\def\mV{{\bm{V}}}

\DeclareMathAlphabet{\mathsfit}{\encodingdefault}{\sfdefault}{m}{sl}
\SetMathAlphabet{\mathsfit}{bold}{\encodingdefault}{\sfdefault}{bx}{n}

\def\gG{{\mathcal{G}}}

\title{Training-Free Multimodal Deepfake Detection via Graph Reasoning}

\name{Yuxin Liu$^1$, Fei Wang$^{2,3,*}$, Kun Li$^{4}$, Yiqi Nie$^{3}$, Junjie Chen$^{3}$, Yanyan Wei$^{2}$, Zhangling Duan$^{3,*}$, Zhaohong Jia$^1$\thanks{* Corresponding Author: Fei Wang, Zhangling Duan}}
\address{\small{$^1$ School of Internet, Anhui University, Hefei, China}\\
\small{$^2$ School of Computer Science and Information Engineering, Hefei University of Technology, Hefei, China}\\
\small{$^3$ Institute of Artificial Intelligence, Hefei Comprehensive National Science Center, Hefei, China}\\
\small{$^4$ Department of Computer Science, Hong Kong Baptist University, Hong Kong, China }
}
\begin{document}
\ninept
\maketitle
\begin{abstract}
Multimodal deepfake detection (MDD) aims to uncover manipulations across visual, textual, and auditory modalities, thereby reinforcing the reliability of modern information systems. 
Although large vision-language models (LVLMs) exhibit strong multimodal reasoning, their effectiveness in MDD is limited by challenges in capturing subtle forgery cues, resolving cross-modal inconsistencies, and performing task-aligned retrieval.
To this end, we propose Guided Adaptive Scorer and Propagation In-Context Learning (GASP-ICL), a training-free framework for MDD.
GASP-ICL employs a pipeline to preserve semantic relevance while injecting task-aware knowledge into LVLMs.
We leverage an MDD-adapted feature extractor to retrieve aligned image-text pairs and build a candidate set.
We further design the Graph-Structured Taylor Adaptive Scorer (GSTAS) to capture cross-sample relations and propagate query-aligned signals, producing discriminative exemplars.
This enables precise selection of semantically aligned, task-relevant demonstrations, enhancing LVLMs for robust MDD.
Experiments on four forgery types show that GASP-ICL surpasses strong baselines, delivering gains without LVLM fine-tuning.

\end{abstract}


\section{Introduction}
\label{sec:intro}
Multimodal deepfake detection (MDD) focuses on identifying manipulated content by jointly modeling visual, textual, and auditory modalities~\cite{ABSTRACT1,ABSTRACT2}, thereby serving as a cornerstone for enhancing the reliability and trustworthiness of modern information systems~\cite{intrro_import,zhao2025temporal,wang2025exploiting,chen2024interclip,chen2025seeing}.
The key challenge, however, lies in effectively capturing subtle forgery cues that are dispersed across modalities~\cite{intrro_subtil,wang2025task}, as well as in resolving the inherent inconsistencies that arise between them.
With the rapid progress of MDD, recent research has shifted toward developing solutions that emphasize stronger generalization and enhanced robustness.
Yu~\etal~\cite{yu2025unlocking} proposed a framework that combines knowledge-guided feature decomposition and forgery prompt learning, aligning image-text embeddings for forgery detection and localization while generating fine-grained prompts to highlight suspicious regions for LLM inference.
However, it heavily relies on the quality of specific forgery descriptions and prompts, and its generalization capability across diverse forgery scenarios remains insufficient.
Sun~\etal~\cite{sun2025towards} annotation workflow, which generates more precise synthetic text descriptions via synthetic masks and prompts.
However, they exhibit a reliance on high-quality annotations.
Although LVLMs show remarkable potential in multimodal understanding, fine-tuning approaches often suffer from substantial computational overhead and limited generalization across tasks~\cite{Funtuned_qppoch}.
\begin{figure}[t!]
\begin{center}
\includegraphics[width=1\linewidth]{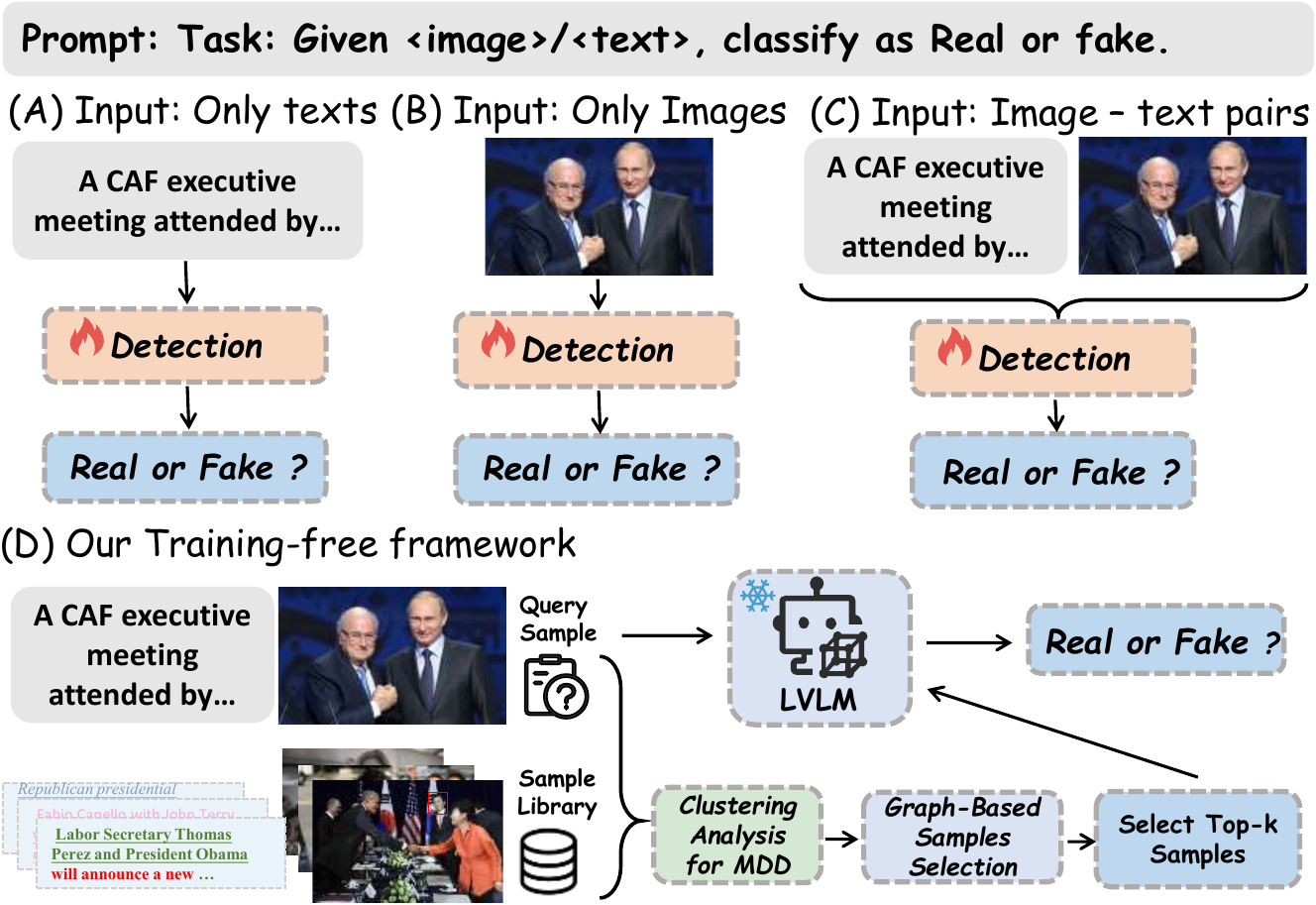}
\vspace{-1.0em}
\caption{
Existing methods: (A) text-only, (B) image-only, (C) image-text paired.
Our method: (D) a training-free multimodal framework that integrates in-context learning with graph reasoning, adaptively mining aligned, task-relevant demonstrations to capture subtle cross-modal forgery cues and enhance robustness across LVLMs.
}
\label{fig:overall}
\end{center}
\vspace{-2.0em}
\end{figure}

In-Context Learning (ICL)~\cite{ICL1}, as a widely adopted paradigm, typically relies on semantically relevant demonstrations. Effectively harnessing the ICL capabilities of LVLMs for the task of MDD remains a challenging objective.
To this end, our research focuses on the following challenges:
(i) Training LVLMs typically requires substantial computational resources, and applying MDD to training-free LVLMs based on ICL faces the difficulty of ensuring robust generalization across diverse forgery types and scenarios.
(ii) Directly applying MDD to training-free LVLMs based on ICL often fails to capture subtle tampering traces and cross-modal inconsistencies.
(iii) The essence of ICL lies in retrieving high-quality samples that are most relevant to the downstream task as prompts.
Yet, simple similarity-based ranking struggles to discriminate fine-grained differences between genuine and forged samples and to achieve precise semantic and structural alignment between queries and exemplars.

To address these challenges, we propose \textbf{Guided Adaptive Scorer and Propagation In-Context Learning (GASP-ICL)}, a training-free ICL framework that leverages a structured exemplar selection pipeline.
(1) In the first stage, we compute the joint similarity between images and text within the feature representation space optimized for MDD, thereby providing discriminative contextual information to support subsequent example selection and inference.
(2) In the second stage, we introduce a graph-based adaptive scorer named \textbf{Graph-
Structured Taylor Adaptive Score (GSTAS)} to evaluate candidate examples. 
This scorer leverages graph structure modeling to capture semantic and structural relationships across samples explicitly, and dynamically amplifies query-aligned nodes through adaptive propagation and a Taylor gate mechanism, thereby surfacing latent manipulation cues.
The resulting high-relevance exemplars serve as discriminative contextual signals to guide LVLM reasoning, enhancing contextual understanding and task alignment without requiring additional training.
As shown in Figure~\ref{fig:overall}, existing methods mainly rely on unimodal text, image, or simple multimodal fusion, and typically require additional training.
In contrast, our proposed framework explicitly integrates exemplar quality and contextual relevance into the ICL process without training, enabling more robust detection of subtle cross-modal inconsistencies in MDD.
Overall, our contributions are as follows:
\begin{itemize}[noitemsep, topsep=0pt, leftmargin=10 pt]

\item We propose \textbf{GASP-ICL}, a training-free framework that leverages task-relevant few-shot examples to construct discriminative contexts, enhancing LVLM-based multimodal deepfake detection.  

\item We introduce a structured pipeline that computes joint image-text similarity in an MDD-oriented feature space, providing effective cues for demonstration selection.  

\item We design a \textbf{GSTAS}, which models cross-sample relations and adaptively propagates query-aligned signals via Taylor gating, yielding highly discriminative exemplars.  

\item Extensive experiments verify that GASP-ICL significantly improves MDD performance and generalizes well across diverse forgery types and complex scenarios.  
\end{itemize}

\begin{figure*}[t!]
\begin{center}
\includegraphics[width=1.0\linewidth]{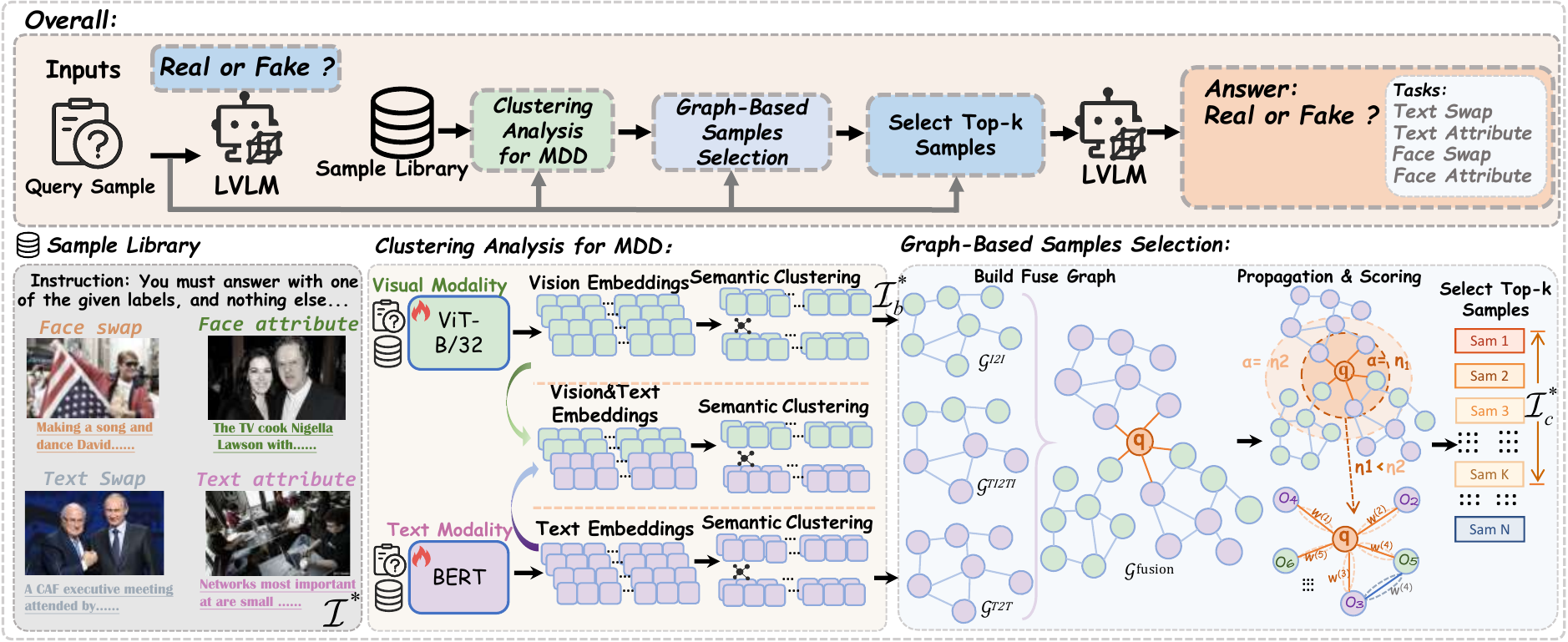}
\vspace{-1.5em}
\caption{Overview of the training-free framework.
(1) Within an MDD feature space, compute joint image-text similarity and extract vision, text, and joint embeddings for semantic clustering.
(2) Subgraphs are constructed, fused, and adaptively scored to select task-relevant samples.
}
\label{fig:method}
\end{center}
\vspace{-1.5em}
\end{figure*}

\section{Methodology} \label{sec:format}

\subsection{Overview}\label{sec:overview}
This work focuses on MDD by leveraging LVLMs without task-specific fine-tuning. We formulate MDD as a binary classification problem over multimodal inputs, where each sample consists of a visual input $I \in \mathbb{R}^{H \times W \times 3}$ and a textual input $T \in \mathbb{R}^L$. The model predicts whether the input is manipulated or authentic.
In the standard ICL pipeline, a frozen LVLM $\mathcal{L}(\cdot)$ is provided with an augmented prompt $\mathcal{P}$ that integrates the query pair $(I, T)$ with multimodal demonstrations sampled from a candidate set $\mathcal{I}^* = \{(I_i, T_i)\}_{i=1}^{N}$, where $N$ denotes the total number of candidates.
The model then generates the prediction $\mathcal{Y}_\textrm{ICL}$ as:
\begin{equation}
\mathcal{Y}_\textrm{ICL} = \mathcal{L}([\mathcal{P}; I, T]).
\end{equation}

However, previous ICL methods largely relied on similarity-based retrieval and thus failed to capture the subtle cross-modal inconsistencies that characterize multimodal deepfakes. 
To address this limitation, we propose GASP-ICL, a training-free framework that adopts a structured selection pipeline to construct compact, task-driven prompts tailored to each query, as illustrated in Figure~\ref{fig:method}.
(1) We first encode the query sample $(I, T)$ and all knowledge base entries 
$\mathcal{I}^* = \{(I_i, T_i)\}_{i=1}^N$ into a joint multimodal embedding space, 
which is specifically adapted to the MDD task to capture subtle forgery cues and cross-modal inconsistencies.
Based on image-text similarity, we retrieve the top-$k_1$ semantically aligned candidates, yielding a set $\mathcal{I}_{b}^*=\{(I_i,T_i)\}_{i=1}^{k_1}\subset \mathcal{I}^*$.
(2) On top of $\mathcal{I}_{b}^*$, we construct a unified fusion graph where edges are defined by similarity in the embedding space, capturing cross-sample semantic and structural relationships. 
We then apply our proposed GSTAS to evaluate the nodes on this graph, assigning discriminative relevance scores that emphasize query-consistent samples.
This process selects the top-$k_2$ most informative demonstrations, resulting in the final demonstration set $\mathcal{I}_{c}^* = \{(I_i, T_i)\}_{i=1}^{k_2} \subset \mathcal{I}_{b}^*$.
Therefore, we treat $\mathcal{I}_{c}^*$ as a knowledge summary and form a structured prompt $\mathcal{P}_c$ that provides task-aligned context to the LVLM.
The final prediction $\hat{\mathcal{Y}}_{\text{ours}}$ is obtained by concatenating $\mathcal{P}_c$ with the query sample:
\begin{equation}\label{eq:prompt}
\begin{aligned}
\mathcal{P}_c = \text{Prompt}\big(\sum_{k=1}^{k_2} \mathcal{I}_{c}^*\big),
~~\hat{\mathcal{Y}}_{\text{ours}} = \mathcal{L}\big([\mathcal{P}_c : I, T]\big).
\end{aligned}
\end{equation}

\subsection{Similarity-Based Retrieval in Multimodal Space}
Inspired by recent retrieval-augmented prompting strategies~\cite{kang2024retrieval}, we embed both the query sample $(I, T)$ and all candidate demonstrations $\mathcal{I}^* = \{(I_i, T_i)\}$ into a shared multimodal embedding space. We adopt CLIP encoders~\cite{clip}, $\mathcal{E}_v(\cdot)$ and $\mathcal{E}_t(\cdot)$, which are further fine-tuned on DGM$^4$~\cite{Shao2023DGM4}, to obtain modality-specific representations $\mathcal{E}_v(I), \mathcal{E}_t(T)$ and $\{\mathcal{E}_v(I_i), \mathcal{E}_t(T_i)\}$.  

We define the retrieval mode as $M \in {\textrm{I2I}, \textrm{T2T}, \textrm{TI2TI}}$, where similarity is computed in three feature spaces: in the visual space (I2I), $\mathcal{O}_{\textrm{I2I}}(i) = \textrm{sim}(\mathcal{E}_v(I), \mathcal{E}_v(I_i))$; in the textual space (T2T), $\mathcal{O}_{\textrm{T2T}}(i) = \textrm{sim}(\mathcal{E}_t(T), \mathcal{E}_t(T_i))$; and in the joint multimodal space (TI2TI), $\mathcal{O}_{\textrm{TI2TI}}(i) = \textrm{sim}(\mathcal{E}_v(I)\oplus \mathcal{E}_t(T), \, \mathcal{E}_v(I_i)\oplus \mathcal{E}_t(T_i))$. 

According to the chosen retrieval mode $M$, the top-$k_1$ candidates are retained to form the coarse candidate set:
\begin{equation}
\mathcal{I}_{b}^* = \operatorname{Top}\text{-}k_1\big(\mathcal{O}_{M}(i)\big).
\end{equation}

\subsection{Graph Construction and Query-Centric Fusion}
Graph construction provides a structured context for ICL~\cite{kang2024retrieval}, enabling LVLMs to better capture cross-modal inconsistencies and subtle forgery cues in MDD.
Each element in $\mathcal{I}_{b}^*$ is treated as a node in $\mV^M$, and edges $\mE^M$ are established by connecting each node to its top-$k_e$ neighbors according to the similarity score $S_M(\cdot)$. For each retrieval mode $M$, the graph is constructed in its corresponding feature space, yielding
\begin{equation}
\gG^{M} = (\mV^M, \mE^M).
\end{equation}

To facilitate ICL in MDD, we construct a query-centric fused graph that unifies heterogeneous evidence and captures cross-modal inconsistencies for robust detection.
Thus, we adopt a query-centric fusion strategy. 
For a given query sample, denoted as the query node $V_q$, its neighborhoods across different modality-specific graphs are aligned and aggregated, while edges are re-weighted by modality-specific coefficients $\lambda_M$ to ensure balanced contributions.
This process yields the fused graph:
\begin{equation}
\gG^\textrm{fusion} = (\mV^\textrm{fusion}, \mE^\textrm{fusion}),
\end{equation}
\begin{equation}\label{eq:fusion_graph}
\begin{aligned}
\mV^\textrm{fusion} &= \big(\bigcup_M \mV^M \big) \cup V_q, \\
\mE^\textrm{fusion} &= \big(\sum_M \lambda_M \mE^M \big) \cup \mE^q,
\end{aligned}
\end{equation}
where $\mE^q$ denotes the additional connections anchored at the $V_q$.

\subsection{Graph-Structured Taylor Adaptive Scorer}
To obtain task-aligned ICL scores for MDD, we propose the \textbf{GSTAS}, which propagates the query’s activation over the fused graph and applies a Taylor-expanded gating mechanism to up-weight manipulation-consistent nodes while suppressing artifacts.

The propagation state is initialized as the standard basis vector $p^{(0)}\in \mathbb{R}^{|\mV^\textrm{fusion}|}$, with unit activation at the query node $V_q$ and zeros elsewhere.
Building on the adjacency construction in ~\eqref{eq:fusion_graph}, we define the fused propagation operator $\mP$ as the normalized adjacency matrix obtained by aggregating modality-specific adjacencies and adding the connection between the query and its anchors.
At the propagation step $t$, the propagation state $p^{(t)}$ is updated over the graph by the operator $\mP$:
\begin{equation}
p^{(t)} = \mP\, p^{(t-1)}.
\end{equation}

Given the node embeddings $\{\mathcal{E}_i\}_{i=1}^{|\mV^\textrm{fusion}|}$, the features at step $t$ are aggregated using a probability weighted sum:
At step $t$, features are aggregated by probability weighting:
\begin{equation}
e^{(t)} = \sum_{i=1}^{{|\mV^\textrm{fusion}|}} p^{(t)}_{i}\, \mathcal{E}_i .
\end{equation}

Here, $p^{(t)}_{i}$ is the $i$-th entry of $p^{(t)}$ and $\mathcal{E}_i$ is the embedding of node $i$, which can be instantiated in the visual space $\mathcal{E}_v(\cdot)$, the textual space $\mathcal{E}_t(\cdot)$, or their joint representation $\mathcal{E}_v(\cdot)\oplus\mathcal{E}_t(\cdot)$ as defined by the retrieval similarities.
To adaptively emphasize query-relevant nodes at propagation step $t$, we introduce a geometric gating weight $w^{(t)}$:
\begin{equation}
w^{(t)} = (1-\alpha\,e^{(t)})^{-1} - 1,
\end{equation}
where $\alpha \in (0,1]$ controls the effective propagation range with $|\alpha\,e^{(k)}|<1$.
Under this condition, the geometric weight admits an infinite Taylor expansion~\cite{wang2024eulermormer,wang2024frequency}, given by
\begin{equation}
w^{(k)} = \sum_{n=0}^{\infty} (\alpha\,e^{(k)})^n .
\end{equation}

A small $\alpha$ rapidly attenuates propagation around the query, while values close to $1$ retain high-order terms and enable long-range information flow.
For each candidate node $i \in \mV^{\textrm{fusion}}$, we aggregate stepwise contributions into a final score $\mathcal{O}(q,i)$ as:
\begin{equation}
\mathcal{O}(q,i) \;=\; \sum_{t=1}^{T}\, w^{(t)}\, p_{i}^{(t)},
\end{equation}
where $T$ is the total number of steps.
We rank all candidate nodes by their scores in descending order and select the top-$k_2$ exemplars:
\begin{equation}
\mathcal{I}_{c}^* = \operatorname{Top}\text{-}k_2\big(\mathcal{O}(i)\big),
\end{equation}
thereby defining the final task-aligned prompt set in ~\eqref{eq:prompt}.

This approach enhances ICL for MDD by scoring candidates that preserve semantic alignment while exposing subtle cross-modal inconsistencies, thereby improving the model’s capacity to address the core challenges of the task.

\section{EXPERIMENT} \label{sec:format}
\subsection{Experiment Setup}\label{DAT}
\subsubsection{\bfseries Datasets \& Evaluation Metrics}
We evaluate our approach on DGM$^4$~\cite{Shao2023DGM4}, using four basic manipulation types: text swap, face swap, text attribute, and face attribute. 
To ensure efficiency and fair comparison, we construct a dedicated sample library for each manipulation type, containing 500 forged and 500 original samples. 
Each sample is an image-text pair with abuse labels under a unified evaluation protocol. We report performance in terms of F1 Score (F1\%) and Accuracy (Acc.\%).

\subsubsection{\bfseries Implementation Details}
In our experiments, we construct a static candidate set as the knowledge base $\mathcal{I}^*$ for each model by randomly sampling $N$=100 image-text pairs from the DGM$^4$ training set.
Following the GASP-ICL pipeline, the first stage retrieves the top-$k_1$ candidates by computing joint image-text similarity in a CLIP fine-tuned feature space for MDD (with $k_1{=}50$), and the second stage applies GSTAS to score them and surface the most discriminative $k_2$ exemplars,
corresponding to one-shot, two-shot, and three-shot settings with $k_2{=}1,2,3$.
To comprehensively assess the effectiveness of our method, we validate our method on seven representative LVLMs, including Qwen2.5-VL-7B~\cite{qwenvl}, InternVL3-8B~\cite{zhu2025internvl3}, Gemma-3-12B~\cite{team2025gemma}, LlaVa-v1.6-7B~\cite{llava3}, Janus-Pro-7B~\cite{Janus-pro}, Owl2.1-7B~\cite{mplug-owl2}, and Kimi-VL-16B~\cite{kimivl}, under identical few-shot settings, reporting results across all manipulation types.
We further connect the subgraphs to the query node $V_q$ using fused edge weights ($\lambda_\textrm{I2I}{=}0.3$, $\lambda_\textrm{I2I}{=}0.4$, $\lambda_\textrm{IT2IT}{=}0.3$) and conduct ablation studies on the propagation range factor $\alpha$ in GSTAS.
In addition, we evaluate the same set of LVLMs under three configurations: zero-shot inference, frozen CLIP, and fine-tuned CLIP.
All models are deployed with vLLM~\cite{vllm}.

\subsection{Results and Comparison}\label{DAT}
\noindent{\bfseries 1) Zero-shot Evaluation on Manipulation Categories.} 
We compare GASP-ICL with two representative paradigms:
Vanilla zero-shot inference and our method with three-shot achieves the best performance.
As shown in Table~\ref{tab:vlm-fewshot}, GASP-ICL consistently improves performance across seven LVLMs and four challenging forgery types, whereas vanilla zero-shot inference remains unstable without task-aware guidance to detect subtle cross-modal cues.
Notably, \textbf{Qwen2.5-VL} achieves the best overall performance across all four forgery types, clearly outperforming other LVLMs.

\begin{table}[t!]
\caption{Comparison between seven LVLMs the proposed GASP-ICL (\textbf{Ours}) in terms of Acc (\%) and F1 score (\%) on DGM$^4$.  
}
\centering
\setlength{\tabcolsep}{3pt}
\renewcommand{\arraystretch}{1.05}
\resizebox{1.0\linewidth}{!}{
\begin{tabular}{|c|cccccccc|cc|}
\hline\thickhline
\multirow{2}{*}{\textbf{Methods}} 
& \multicolumn{2}{c}{\textbf{Face Swap}} 
& \multicolumn{2}{c}{\textbf{Face Attribute}} 
& \multicolumn{2}{c}{\textbf{Text Swap}} 
& \multicolumn{2}{c|}{\textbf{Text Attribute}} 
& \multicolumn{2}{c|}{\textbf{Overall}} \\
& \textbf{Acc} & \textbf{F1}  
& \textbf{Acc} & \textbf{F1}  
& \textbf{Acc} & \textbf{F1} 
& \textbf{Acc} & \textbf{F1} 
& \textbf{Acc} & \textbf{F1}  \\
\hline 
InternVL3~\cite{zhu2025internvl3} &44.24&39.12&43.61&40.25&42.82&39.64&43.10&39.93&43.44&39.74\\
\rowcolor{gray!20} \textbf{+ Ours} &\textbf{47.80}&\textbf{45.32}&\textbf{46.90}&\textbf{47.53}&\textbf{45.86}&\textbf{46.82}&\textbf{46.54}&\textbf{46.92}&\textbf{46.78}&\textbf{46.65}
\\ \hline
Gemma-3~\cite{team2025gemma} &46.20&41.50&47.11&42.33&45.26&41.00&45.86&41.63&46.11&41.62\\
\rowcolor{gray!20} \textbf{+ Ours}  &\textbf{49.69}&\textbf{47.83}&\textbf{49.19}&\textbf{49.32}&\textbf{47.20}&\textbf{48.17}&\textbf{47.82}&\textbf{48.92}&\textbf{47.73}&\textbf{48.56}
\\ \hline
LLaVA-v1.6~\cite{llava3}
&43.18&38.49&44.00&39.25&42.32&38.76&42.91&39.10&43.10&38.90\\
\rowcolor{gray!20} \textbf{+ Ours} &\textbf{46.20}&\textbf{44.13}&\textbf{46.80}&\textbf{46.94}&\textbf{45.12}&\textbf{45.90}&\textbf{45.71}&\textbf{46.27}&\textbf{45.96}&\textbf{45.81}\\ \hline
Janus-Pro~\cite{Janus-pro} &40.17&38.26&39.45&40.63&39.94&41.91&42.58&42.47&40.54&40.82\\
\rowcolor{gray!20} \textbf{+ Ours} &\textbf{43.91}&\textbf{41.23}&\textbf{45.74}&\textbf{47.02}&\textbf{48.67}&\textbf{49.46}&\textbf{43.54}&\textbf{44.32}&\textbf{45.47}&\textbf{45.51}\\ \hline
Owl2.1~\cite{mplug-owl2} &44.53&39.39&45.24&40.15&43.45&39.29&44.10&39.72&44.33&39.64\\ 
\rowcolor{gray!20} \textbf{+ Ours} &\textbf{47.76}&\textbf{44.93}&\textbf{46.47}&\textbf{47.15}&\textbf{45.76}&\textbf{46.24}&\textbf{45.62}&\textbf{46.54}&\textbf{46.40}&\textbf{46.21}\\  \hline
Kimi-VL~\cite{kimivl} &49.65&42.38&51.63&43.33&49.54&40.83&50.11&38.62&50.23&41.29\\ 
\rowcolor{gray!20}  \textbf{+ Ours} &\textbf{53.85}&\textbf{49.29}&\textbf{51.29}&\textbf{45.65}&\textbf{47.83}&\textbf{48.53}&\textbf{50.47}&\textbf{43.52}&\textbf{50.86}&\textbf{46.74}\\  \hline
Qwen2.5-VL~\cite{qwenvl}
&47.50&38.58&50.80&38.31&50.84&38.34&49.70&35.82&49.71&37.76\\
\rowcolor{gray!20}  \textbf{+ Ours}       &\textbf{54.30}&\textbf{54.29}&\textbf{53.20}&\textbf{52.65}&\textbf{52.40}&\textbf{46.58}&\textbf{52.30}&\textbf{50.62}&\textbf{53.05}&\textbf{51.04}\\
\hline
\end{tabular}}
\label{tab:vlm-fewshot}
\vspace{-0.5em}
\end{table}

\begin{table}[t!]
\caption{Performance ablation of Top-$k_2$ in the proposed GSTAS.}
\centering
\setlength{\tabcolsep}{4.5pt}
\renewcommand{\arraystretch}{1.05}
\resizebox{1.0\linewidth}{!}{
\begin{tabular}{|c|cccccccc|cc|}
\hline\thickhline
\multirow{2}{*}{\textbf{Setting}} &  
\multicolumn{2}{c}{\textbf{Face Swap}} & 
\multicolumn{2}{c}{\textbf{Face Attribute}} &
\multicolumn{2}{c}{\textbf{Text Swap}} &
\multicolumn{2}{c|}{\textbf{Text Attribute}}  &
\multicolumn{2}{c|}{\textbf{Overall}} \\
& \textbf{Acc} & \textbf{F1} 
& \textbf{Acc} & \textbf{F1}  
& \textbf{Acc} & \textbf{F1}  
& \textbf{Acc} & \textbf{F1}  
& \textbf{Acc} & \textbf{F1}  \\
\hline
Top-1 &53.90&51.78&52.20&49.48&51.60&45.95&50.50&45.20&52.05&48.10\\
Top-2 &54.10&51.83&52.80&51.36&52.10&46.30&51.00&50.25&52.50&49.94\\
\rowcolor{gray!20} \textbf{Top-3} &\textbf{54.30}&\textbf{54.29}&\textbf{53.20}&\textbf{52.65}&\textbf{52.40}&\textbf{46.58}&\textbf{52.30}&\textbf{50.62}&\textbf{53.05}&\textbf{51.04}\\
\hline
\end{tabular}
}
\label{tab:qwen-fewshot}
\vspace{-0.5em}
\end{table}
\begin{table}[t!]
\caption{GSTAS performance with varying $\alpha$ on manipulation types.}
\centering
\setlength{\tabcolsep}{4.5pt}
\renewcommand{\arraystretch}{1.05}
\resizebox{1.0\linewidth}{!}{
\begin{tabular}{|c|cccccccc|cc|}
\hline\thickhline
\multirow{2}{*}{\textbf{Setting}} 
& \multicolumn{2}{c}{\textbf{Face Swap}} 
& \multicolumn{2}{c}{\textbf{Face Attribute}} 
& \multicolumn{2}{c}{\textbf{Text Swap}} 
& \multicolumn{2}{c|}{\textbf{Text Attribute}}
& \multicolumn{2}{c|}{\textbf{Overall}} \\
& \textbf{Acc} & \textbf{F1} 
& \textbf{Acc} & \textbf{F1} 
& \textbf{Acc} & \textbf{F1} 
& \textbf{Acc} & \textbf{F1} 
& \textbf{Acc} & \textbf{F1}  \\
\hline
\textbf{$\alpha$} = 0.2 
&50.67&49.29&50.43&48.29&49.93&44.38&50.13&48.97&50.29&47.73\\
\rowcolor{gray!20}\textbf{$\alpha$ = 0.4 }
&\textbf{54.30}&\textbf{54.29}&\textbf{53.20}&\textbf{52.65}&\textbf{52.40}&\textbf{46.58}&\textbf{52.30}&\textbf{50.62}&\textbf{53.05}&\textbf{51.04}\\
\textbf{$\alpha$} = 0.6 
&54.07&53.39&52.94&50.86&51.74&45.78&51.60&49.69&52.58&49.93\\
\textbf{$\alpha$} = 0.8     
&53.93&49.75&51.48&49.98&50.71&44.64&50.02&49.94&51.54&48.58\\
\textbf{$\alpha$} = 1.0 
&53.25&49.34&50.39&48.41&50.09&44.51&49.87&49.18&50.90&47.86\\
\bottomrule
\end{tabular}
}
\label{tab:GSTAS-alpha}
\vspace{-0.5em}
\end{table}
\begin{table}[t!]
\caption{Performance of GSTAS under three configurations: zero-shot inference, frozen CLIP ($\odot$), and fine-tuned CLIP ($\otimes$).}
\centering
\setlength{\tabcolsep}{4.5pt}
\renewcommand{\arraystretch}{1.05}
\resizebox{\linewidth}{!}{
\begin{tabular}{|c|cccccccc|cc|}
\hline\thickhline
\multirow{2}{*}{\textbf{Setting}} 
& \multicolumn{2}{c}{\textbf{Face Swap}} 
& \multicolumn{2}{c}{\textbf{Face Attribute}} 
& \multicolumn{2}{c}{\textbf{Text Swap}} 
& \multicolumn{2}{c|}{\textbf{Text Attribute}} 
& \multicolumn{2}{c|}{\textbf{Overall}} \\
& \textbf{Acc} & \textbf{F1} 
& \textbf{Acc} & \textbf{F1} 
& \textbf{Acc} & \textbf{F1} 
& \textbf{Acc} & \textbf{F1} 
& \textbf{Acc} & \textbf{F1}  \\
\hline 
Zero-Shot
&47.50&38.58&50.80&38.31&50.84&38.34&49.70&35.82&49.71&37.76\\
$\odot$CLIP
&53.18&52.72&52.98&50.73&52.27&46.43&51.16&49.43&52.40&49.90\\
\rowcolor{gray!20}\textbf{$\otimes$CLIP}&\textbf{54.30}&\textbf{54.29}&\textbf{53.20}&\textbf{52.65}&\textbf{52.40}&\textbf{46.58}&\textbf{52.30}&\textbf{50.62}&\textbf{53.05}&\textbf{51.04}\\
\hline
\end{tabular}
}
\label{tab:vlm-init-tuned}
\vspace{-0.5em}
\end{table}

\begin{figure}[t!]
\begin{center}
\includegraphics[width=1\linewidth]{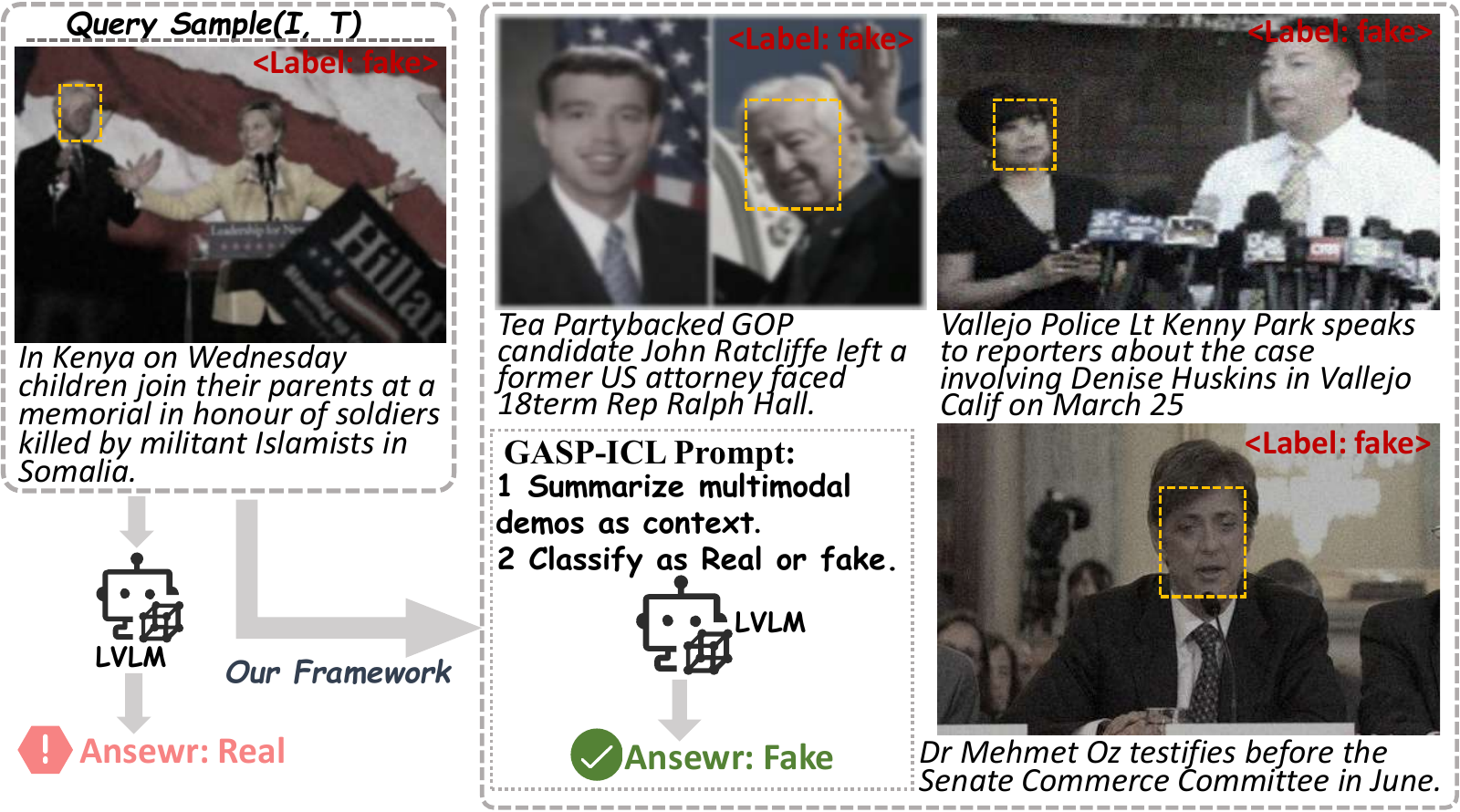}
\vspace{-1.0em}
\caption{Case study of GASP-ICL on Face Swap. Direct LVLM inference misclassifies, while guidance distilled from the top-3 samples leads to the correct result.
}
\label{fig:case}
\end{center}
\vspace{-2.0em}
\end{figure}

\noindent{\bfseries 2) Few-shot Performance Under Different Settings.}
Table~\ref{tab:qwen-fewshot} reports the few-shot performance of Qwen2.5-VL under different shot settings across four manipulation types, evaluated under identical experimental conditions. 
The results indicate that three-shot achieves the best performance within the model’s context window, as it provides sufficient task-aware demonstrations to capture the subtle cross-modal forgery cues in MDD, while avoiding the noise and staying within the model’s context window limit.

\noindent{\bfseries 3) Few-shot Performance Under Different Settings.}
Figure~\ref{fig:case} presents a challenging sample where direct zero-shot inference results in an incorrect prediction. Instead, GASP-ICL leverages a structured pipeline to select high-quality demonstrations, each guiding the model to the correct prediction. 
These results highlight the strength of our method in eliminating misleading samples and generating task-aligned prompts that enhance model reasoning.

\noindent{\bfseries 4) Impact of the propagation range factor $\alpha$ in GSTAS.} 
We vary $\alpha \in \{0.2,0.4,0.6,0.8,1\}$ to study its effect on propagation.
Table~\ref{tab:GSTAS-alpha} shows that a small $\alpha$ restricts propagation in GSTAS and hinders capturing subtle cross-modal forgery cues, whereas a large $\alpha$ causes over-propagation, introduces noise, and degrades exemplar discriminability.
Among all settings, $\alpha=0.4$ achieves the best balance, resulting in the highest overall performance.

\noindent{\bfseries 5) Impact of task-specific CLIP adaptation on GSTAS.}
As shown in Table~\ref{tab:vlm-init-tuned}, on Qwen2.5-VL, the pretrained CLIP results in lower performance with accuracy $52.40\%$ and F1 $49.90\%$, compared to the task-adapted CLIP in GASP-ICL.
This clearly highlights that task-specific CLIP adaptation substantially enhances cross-modal alignment and improves sensitivity to subtle manipulations.



\section{CONCLUSIONS} \label{sec:format}
This paper introduces GASP-ICL, a novel training-free framework for MDD.
By leveraging a feature extractor adapted for MDD tasks and graph-structured discriminative alignment, GASP-ICL adaptively selects informative and context-relevant demonstrations.
This adaptive scoring strategy effectively guides in-context learning to capture subtle forgery cues from cross-modal inconsistencies.
Extensive experiments on four types of forgeries validate its effectiveness, demonstrating a generalizable and scalable solution for large vision-language models in security-critical applications.
\pagebreak
\bibliographystyle{IEEEbib}
\bibliography{refs}
\end{document}